\documentclass[conference]{IEEEtran}
\IEEEoverridecommandlockouts
\usepackage{cite}
\usepackage{amsmath,amssymb,amsfonts}
\usepackage{algorithmic}
\usepackage{dblfloatfix}
\usepackage{graphicx}
\usepackage{textcomp}
\usepackage{url}
\usepackage{graphicx}
\usepackage{caption}
\usepackage[table]{xcolor}
\def\BibTeX{{\rm B\kern-.05em{\sc i\kern-.025em b}\kern-.08em
    T\kern-.1667em\lower.7ex\hbox{E}\kern-.125emX}}

\begin{document}

\title{Embedded Implementation of a Deep Learning Smile Detector\\
\thanks{This work was partially funded by the Academy of Finland
project 309903 CoefNet. Authors also thank CSC--IT Center for Science for computational resources.}
}

\author{\IEEEauthorblockN{Pedram Ghazi, Antti P. Happonen, Jani Boutellier, and Heikki Huttunen} 
\textit{Tampere University of Technology}\\
Tampere, Finland}

\maketitle

\begin{abstract}
In this paper we study the real time deployment of deep learning algorithms in low resource computational environments. As the use case, we compare the accuracy and speed of neural networks for smile detection using different neural network architectures and their system level implementation on NVidia Jetson embedded platform. We also propose an asynchronous multithreading scheme for parallelizing the pipeline. Within this framework, we experimentally compare thirteen widely used network topologies. The experiments show that low complexity architectures can achieve almost equal performance as larger ones, with a fraction of computation required.
\end{abstract}

\begin{IEEEkeywords}
Deep learning, smile recognition, embedded implementation, Nvidia Jetson
\end{IEEEkeywords}

\section{Introduction}

 Facial expression recognition from real-time image data streams has become a topic of intensive research in image analysis. Due to its popularity especially in entertainment, embedded real-time image detector implementations for smile, age and gender are currently emerging. Contemporary commercial platforms include Amazon Rekognition \cite{a1}, Clarifai Image and Video Recognition \cite{a2}, Google Cloud Vision API \cite{a3}, and Microsoft Azure Face API \cite{a4}. All these platforms, however, are mainly based on cloud computing and thus not suitable for real-time low-latency use cases.
 
 Varying pose, illumination and other imaging conditions affect facial image recognition and make image-based analysis challenging in real-time embedded applications---which makes the topic interesting for research purposes. Deep learning or deep convolutional neural network’s (CNN’s) have become one of the flagship applications of modern machine learning (ML) for image and video analysis \cite{a5}. 
 
 \begin{figure}[t]
\centering
\includegraphics[width=0.24\textwidth]{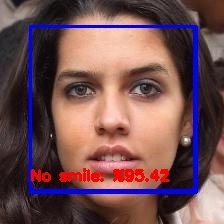}
\includegraphics[width=0.24\textwidth]{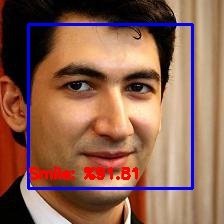}
\caption{Outcome images of the proposed smile detector. An unhappy face with detection accuracy of 95\% (left) and a smiling face with detection accuracy of 92\% (right). The image samples are from ChaLearn Looking At People 2016 dataset.}\label{fig:image1}
\end{figure}

 Since the dawn of deep CNN’s, most researchers have concentrated on increasing the accuracy of deep learning models. Only recently, research focus has started to migrate towards efficient implementations \cite{mobilenets,squeeze} that can run on resource-limited embedded platforms. Moreover, most of the implementations research concentrates on boosting the efficiency of individual components, and only a few system level studies exist (see Section \ref{sec:related}). Yet, complete systems consist of several components with different real time requirements and their interplay has to be carefully designed.
 
 In this paper, we describe a system level architecture of a deep learning algorithm for smile recognition. Smile recognition has several applications, from customer satisfaction measurement \cite{product} to clinical quantiﬁcation of emotional state in patients \cite{tur2018}. Moreover, it serves as an excellent use case for system level architecture design. The full system contains several components whose computational complexities vary: frame grabbing, face localization, smile recognition and visualization. Although the deep neural network may not operate at the full frame rate, the implementation should still appear as real time in the visualized recognition result. Illustration of outcome images of the proposed implementation is shown in Fig. \ref{fig:image1}.
 
\begin{figure*}[!b]
\vspace{1mm}
\resizebox{\textwidth}{!}
{\begin{minipage}{\textwidth}
\centering
\includegraphics[width=0.75\textwidth]{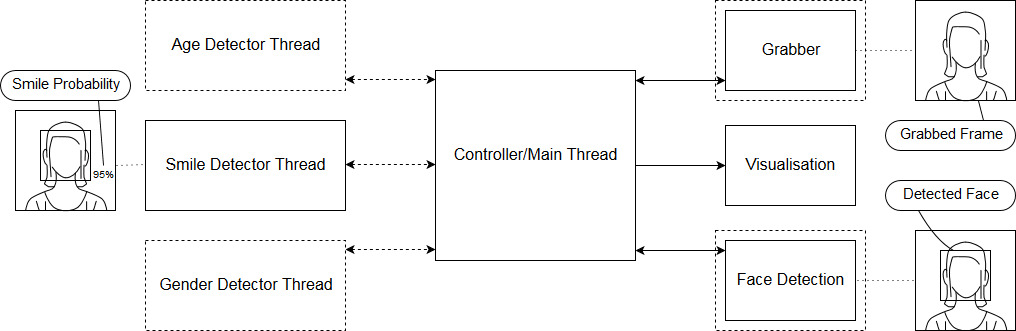}
\end{minipage} }
\caption{Software architecture diagram of the proposed smile detector.}\label{fig:diagram}
\end{figure*}
 
\section{Related Work}
\label{sec:related}

Early algorithms for robust and fast \textit{human and face detection} were published in the beginning of the millennium \cite{dalal2005, viola}. Still today, OpenCV library based on the Viola-Jones algorithm \cite{viola} is routinely employed in many real-time face detection techniques resulting in augmented face region in the image data stream \cite{Cha2014, tsai2015, kazanskiy2017, elrefe2017}.

During the deep learning era, the accuracy of \textit{face detection and recognition} has improved dramatically. Very recently, several variants of CNN based methods for face detection \cite{jiang2017, YangLLT17, Mamalet, Farrugia, Garcia} and recognition \cite{hu2015face, byeon2016, arse2017, ChiZC17} were reported. 

In \textit{facial expression recognition}, the early systems relied on finding the locations of a number of landmark points (eyes, nose, corners of the mouth, etc.), whose relative location was then used as the basis for classification, e.g., between smile and no-smile components, incorporating the applicable criteria that follow. Contemporary deep learning systems feed the raw pixel data of the cropped face area to a deep neural network. This representation is far more informative and enables extraction of information from the entire face area. 

Several studies on detection of apparent age, gender and smile using deep learning methods have been published, e.g. \cite{uri2106}, smile detection for image data of young children \cite{xia2017}, and SmileNet \cite{smilenet} a modern network architecture  for smile detection. A deep multi-task learning framework HyperFace can provide simultaneous face detection, landmarks localization, pose estimation and gender recognition \cite{ranjan2017}.

Nonetheless, one of the challenges is to achieve real-time operation for streaming data in an environment where  system components have different execution times. At the same time, the system should retain a high accuracy, since effective recognition models for \textit{facial expression detection} tend to be computationally expensive. Alongside the object and face detection method using Haar feature-based cascade classifiers \cite{viola}, there exist variety of tailored face detection deep CNN methods for embedded systems. For example, FaceBoxes is a real-time \textit{face detector} with high accuracy using CPU \cite{zhang2017}, whereas LCDet is an 8-bit fixed-point TensorFlow \textit{object detection} on embedded systems \cite{tripathi2017}, and OpenFace provides a general-purpose \textit{face recognition} library with mobile applications \cite{amos2016openface}.

In the next sections, we present a real-time system level architecture of a \textit{smile detector}, which is applicable to embedded solutions for image data streams.

\begin{figure*}
\vspace{1mm}
\resizebox{\textwidth}{!}
{\begin{minipage}{\textwidth}
\centering
\includegraphics[width=0.73\columnwidth]{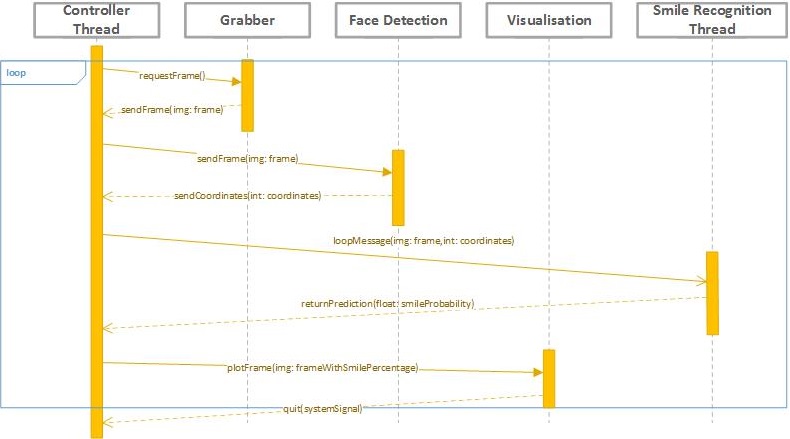}
\end{minipage} }
\caption{Sequence diagram of the proposed system architecture.}\label{fig:sequence}
\end{figure*}

\section{Methods}

\subsection{System Level Architecture}

The target platform for our system implementation is NVidia Jetson TX2, a heterogeneous embedded platform with six ARM compatible CPU cores and an embedded GPU with 256 CUDA cores. For comparison, the experimental section will also benchmark the networks on a Desktop computer with and without a GPU.

For implementing the smile detection system, asynchronous multithreading is used for parallelizing the computation.
Fig. \ref{fig:diagram} illustrates the software architecture of our system. 
The architecture consists of a \textit{main thread} and \textit{worker threads}, each dedicated to one task in the processing pipeline (grabbing, face detection, smile detection, etc.). The threads are created by the main thread and they operate asynchronously by polling for frames from the frame queue held by the main thread. Each grabbed frame holds a binary flag for every thread, indicating whether the frame was already processed by the thread. Every time a thread is polling for a frame, the most recent not yet processed frame is returned (if available). Moreover, the returned frame should meet the necassary prerequisites: for example, the smile detection thread should receive the most recent frame that has passed the face detection stage). After processing, each worker thread will attach the processing result into the frame (for example the face location from the detection thread). 

The modular architecture enables extending the architecture with new features such as \textit{age detection} in the future. These foreseen extensions have been depicted by boxes with a dashed outline. Moreover, the modularity enables easy load balancing between tasks by prioritizing the threads differently, and by multiplying the number of workers for a critical task.

The communication between threads is asynchronous, so the threads do not need to start and finish computation at the same time. Since all threads are working almost at the same pace, the system works in a stable manner and is able to grab and draw frames at 20-25 fps, which is just a little below of the inference rate of smaller networks like Mobilenet. A sequence diagram of thread communications is depicted in Fig. \ref{fig:sequence}.

\subsection{Data Preprocessing}

\subsubsection{Face Detection}

 
Smile detection starts by finding the face location in the image. We rely on the classical, extremely fast Viola-Jones detector \cite{viola}. The deficiency in accuracy is compensated by the real time nature of the detector, and therefore a single missed detection is not essential among a stream of frames. On the other hand, at \textit{training time} the execution speed is not critical, and we use a more thorough face detector \cite{headhunter} in order to take the most out of the annotated training data. The discrepancy of using different methods at training and deployment does not pose a problem, due to the alignment step that compensates for the different behavior of the two detection approaches.

\subsubsection{Face Alignment}

Face alignment is an essential part of our smile detection pipeline. Alignment normalizes the input such that different face parts are always at the same location. This way the network can always trust that for example eyes are always in a fixed location at the input. Moreover, the alignment corrects problems with rotation and scale that might distort the data and degrade the accuracy.

Most alignment approaches in the literature are not designed for real time use. In order to decrease the time consumption, we use a landmark-based approach \cite{dlip}, which finds the location of a fixed number of face landmarks with an ensemble of regression trees and matches the landmarks with a landmark  template set. The preprocessing of the landmark template is done as in our earlier paper \cite{bai2018}.

\begin{figure}
\centering
\includegraphics[width=0.95\columnwidth]{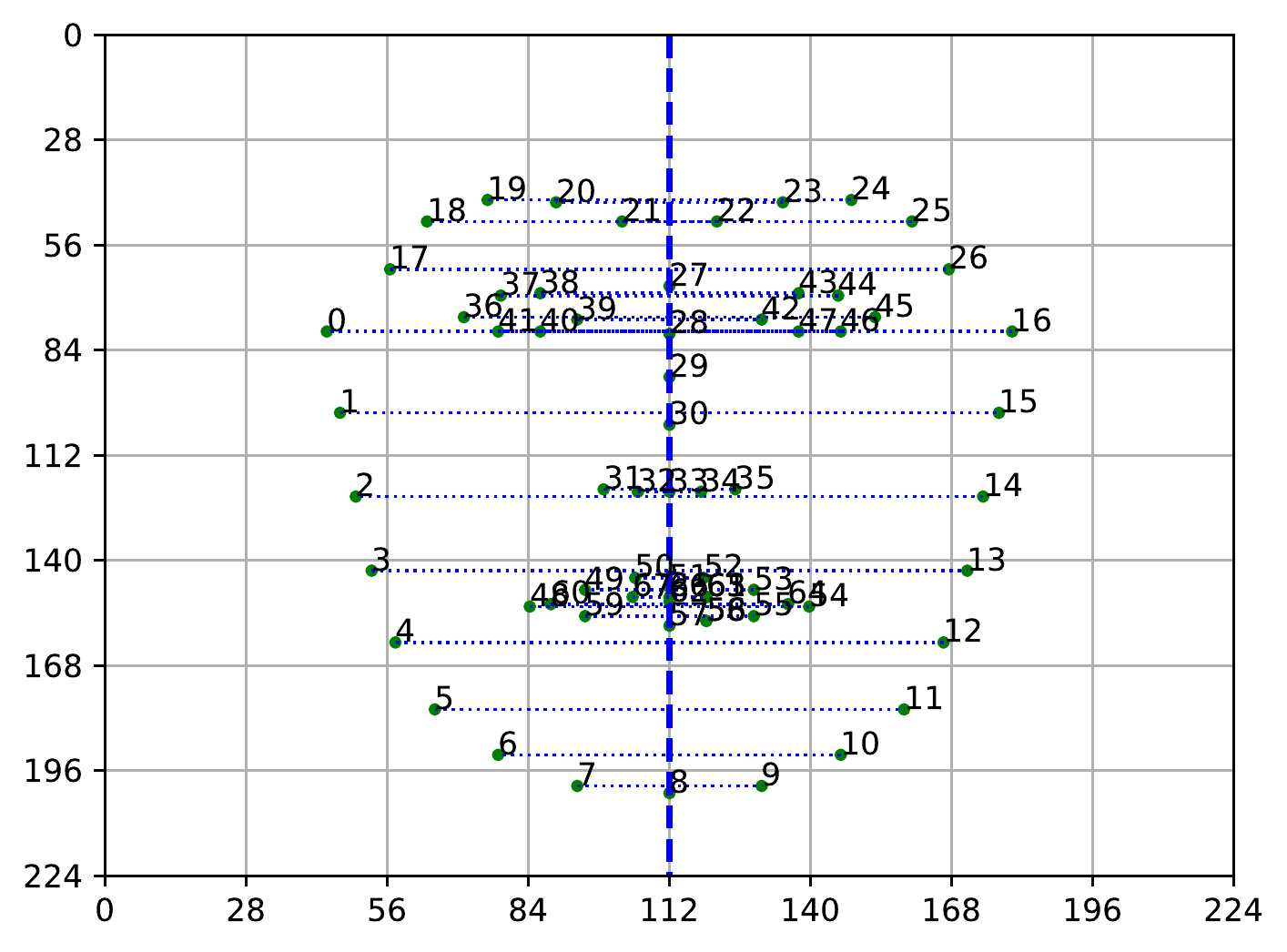}
\caption{Landmark targets for alignment. Symmetric landmarks pairs are highlighted by dashed lines.}\label{fig:landmarks}
\end{figure}

The detected landmarks are matched with a reference landmark set, which was obtained from the landmarks of a randomly selected example from the training data, and normalized to have horizontal symmetry with respect to the centerline of the frame (Figure~\ref{fig:landmarks}). We require a symmetric landmark set in order to enable augmenting the data by left-right-flips of each training sample. 


After detecting the landmarks, we find the best similarity transformation between the two point patterns and map the complete image through the found mapping. A similarity transformation is more limited than a full affine transformation, as it allows only only scaling, rotation and translation (but not shearing). Visually we have discovered, that the affine transformation may sometimes distort the geometry of the face and thus degrade the smile detection accuracy. For a recap of computing the least squares fit of the similarity transformation, we refer the reader to \cite{bai2018}.


\subsection{Network Structures}

We compare six different well-known training architectures: Inception V3 (2014), VGG-16 (2015), Inception-resnet V2 (2016), ResNet (2016), Mobilenet with different paramenters (2017), and XCeption (2017). In particular, we are interested to find a fast, yet accurate architecture, in particular when executed on an embedded platform. The compared architectures are briefly described below.

\textit{Inception V3} \cite{inceptionv3} architecture scales up the network and adds computational efficiency by aggressive regularization and factorized convolutions. Inception V3 architecture provides good performance at relatively low computational cost.

\textit{VGG-16} \cite{vgg} are networks with increasing depth using an architecture with very small (3$\times$3) convolution filters. The depth of these networks can be up to 16--19 weight layers.

In \textit{ResNet-50} \cite{resnet} architecture, the layers are reformulated as learning residual functions with reference to the layer inputs (instead of learning unreferenced functions). ResNet networks are easier to optimize, and the networks can gain accuracy from considerably increased depth.

\textit{Inception-resnet V2} \cite{inc-resnet} architecture with residual connections accelerates the training of Inception networks significantly. Moreover, proper activation scaling stabilizes the training of wide residual Inception networks.

\textit{Mobilenets} \cite{mobilenets} are based on a streamlined architecture that uses depth-wise separable convolutions to build light weight deep neural networks. These networks are designed especially for mobile and embedded vision applications. Adjustable hyper-parameter $\alpha$ and $\rho$ allows the model builder to choose the right sized model for their application based on the constraints of the problem. 
In the experimental section we study several variants of the Mobilenets architecture by choosing the feature map multiplier $\alpha \in \{0.25, 0.5, 0.75, 1.0\}$ and resolution multiplier $\rho \in \{0.714, 1.0\}$ (input resolutions $160\times 160$ and $224\times 224$), respectively.

\textit{XCeption} \cite{xception} is a deep convolutional neural network architecture inspired by Inception. In XCeption architectures, Inception modules have been replaced with depth-wise separable convolutions. XCeption architecture has the same number of parameters as Inception V3, but XCeption networks performance gains are due to a more efficient use of model parameters.

\begin{table}
\centering
\vspace{2mm} 
\begin{center} 
\caption{Computational speed and network size of the compared pre--trained network architectures for smile detection. The network structures are ordered by the number of floating point operations per frame (FLO column). \label{tab:results}}
\end{center}
\centering


\resizebox{\columnwidth}{!}{\begin{tabular}{l||ccccc}

				  			 & FPS				& FPS			   & FPS		   & Size						& FLO    \\
Network Model     			 & (Jetson)			& (CPU)		       & (GPU)	   	   & $\times 10^6$				& $\times 10^9$		\\
\hline\hline
Mobilenet ($\alpha=0.25,\rho=0.714$) 		 & \textbf{28.1}  	& \textbf{31.5}    & \textbf{164}  & \textbf{0.21} 				& \textbf{0.02}\\
Mobilenet ($\alpha=0.25,\rho=1$)		& 27.3  	& 19.6    & 158  & 0.21 				& 0.03\\
Mobilenet ($\alpha=0.5,\rho=0.714$)  	 	 & 26.1		    	& 18.5             & 159           & 0.83						& 0.07\\
Mobilenet ($\alpha=0.5,\rho=1$)   	 & 25.6		    	& 9.0               & 154           & 0.83						& 0.14\\
Mobilenet ($\alpha=0.75,\rho=0.714$) 		 & 24.2				& 12.2             & 153           & 1.8						& 0.16\\
Mobilenet ($\alpha=1,\rho=0.714$)  		 & 22.5          	& 9                & 146.8         & 3.2						& 0.28\\
Mobilenet ($\alpha=0.75,\rho=1$) 		 & 23.5 			& 6.5              & 149           & 1.8						& 0.31\\
XCeption           			 & 9.6            	& 0.9			   & 28            & 20.8						& 0.35\\
Mobilenet ($\alpha=1,\rho=1$)  		 & 22.3           	& 4.9              & 146.1         & 3.2						& 0.55\\
Inception V3      			 & 7.9            	& 2.9              & 22            & 21.8						& 2.8\\
ResNet-50          			 & 8.4            	& 1.5              & 27            & 23.5						& 3.8\\
Inception-resnet V2			 & *              	& 1.5              & 11            & 54.3						& 6.4\\
VGG-16             			 & *              	& 0.9              & 25            & 134						& 15\\
\end{tabular}}


\end{table}

\begin{figure*}
\centering
\includegraphics[width=1\textwidth]{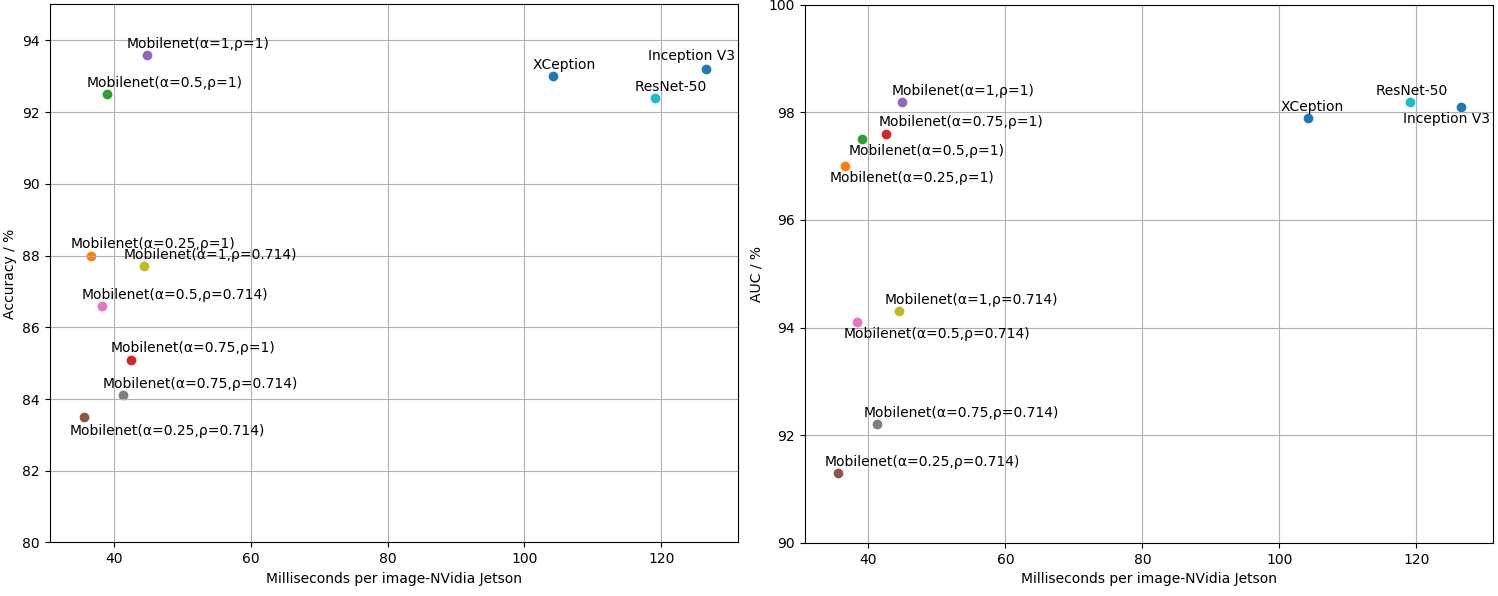}
\captionof{figure}{Accuracy and area under the ROC curve metrics with respect to the prediction time for the  studied network architectures. Results from VGG-16 (ACC: 91.2\%, AUC: 90.5\%) and Inception-resnet-V2 (ACC: 92.8\%, AUC: 98.0\%) models are omitted since their large memory footprint does not allow execution on NVidia Jetson-TX2 embedded platform (both would be outside the image borders to the right). }\label{fig:accuracies}
\end{figure*}

\section{Experiments}

\subsection{Setup}

In our study, we utilized network models trained on ImageNet data such that the last layer of the pre--trained network was removed, and replaced with a dense layer. This was implemented as a binary classification by making use of sigmoid activation layer. 

Also in our study, we applied data augmentation in order to improve the  model generalization. The input images were augmented using shearing, zooming, and  horizontal flipping.

We consider two platforms for measuring the computation speed: a desktop computer with Intel Core i5-6200U CPU and the K40 Desktop GPU and an embedded platform NVidia Jetson TX2.

\subsection{Datasets}

For the experimentation, we use three face datasets: 1) GENKI-4K dataset, 2) CelebFaces Attributes dataset and 3) ChaLearn Looking At People (LAP) 2016 dataset.

The \textit{GENKI-4K dataset}\footnote{\url{http://mplab.ucsd.edu}} includes 4000 image containing expression and head--pose labels. The \textit{CelebA dataset}\footnote{\url{http://mmlab.ie.cuhk.edu.hk/projects/CelebA.html}} is a large--scale face attributes dataset with more than 200 000 images of famous people, each with 40 attribute annotations, including the smile attribute.  
Finally, the \textit{LAP dataset}\footnote{\url{http://chalearnlap.cvc.uab.es/}} is a public dataset of annotated face images originally released for the ICCV2015 looking at people competition. We use the CVPR 2016 competition variant, consisting of 7,591 facial images. Compared to the other datasets, images in the LAP dataset are taken in non--controlled environments and they have more diverse backgrounds. 

\subsection{Evaluation}

We evaluated the computational performance of our smile detector approach employing the thirteen network topologies with the following measures: (1) frames per second (FPS), (2) model size (number of parameters), and (3) the number of floating point operations (FLO). 
 
In addition to the computational speed, we also carried out  accuracy tests with  two quantitative measures: accuracy 
\[
\mathrm{ACC} = \frac{\mathrm{TP} + \mathrm{TN}}{\mathrm{TP} + \mathrm{TN} + \mathrm{FP} + \mathrm{FN}},
\]
and the area under the receiver operating characteristic curve (AUC). ACC and AUC summarize the detection accuracy of the proposed smile detector.




\subsection{Results} 

The speed of the compared networks is summarized in Table~\ref{tab:results}. The network variants are ordered by the number of floating point operations required to process a single frame. One can see that there are huge differences between networks: The heaviest network (VGG16) requires about 750x the number of floating point operations of the lightest (Mobilenet). Although the other performance indicators are closer to each other (due to the parallelization within the processor), the differences are still in the order of a magnitude. Moreover, the model size prohibits the execution of some of the largest models.


In full smile detection pipelines, such as the SmileNet \cite{smilenet}, also a face localization step is included. Thus, in order to evaluate the networks in the context of an entire system, we attach the smile detection network together with the Viola-Jones face detector running on the CPU. In this case, the overall speed on a desktop GPU reaches 40-60 FPS (depending on the input resolution), clearly faster than that of the SmileNet (21.15 FPS on the GENKI-4K dataset \cite{smilenet}). It should be noted, however, that this pipeline does not reach the accuracy shown in Table \ref{tab:comparison}, since the test localization is not at the level of modern CNN based methods. On the other hand, the state of the art methods will not even run on our embedded platform (due to memory footprint inherited from the VGG architecture), let alone in real time.

Nevertheless, when face localization is successful, the smile detection has a high accuracy---even with limited resources. In Figure \ref{fig:accuracies}, the smile detection accuracies are shown on the GENKI-4K dataset. Table \ref{tab:comparison} summarizes the results in comparison with SmileNet on the GENKI-4K and CelebA datasets. The smile detection performance of the proposed method is close to the cutting-edge detectors. Once again, we want to stress that the network used in  \cite{smilenet} will not run on an embedded platform due to its huge memory footprint.

\section{Conclusions}

We have proposed an architecture for real time deployment of deep learning algorithms in low resource computational environments employing NVidia Jetson embedded platform. As an use case, we have implemented a smile detector and evaluated it using thirteen conventional deep neural network models and three popular image databases. 
In particular, we compared the performances using a desktop CPU, NVidia Jetson, and a desktop GPU implementation, and studied the speed/accuracy tradeoff. As a result, we showed that there are significant differences in the computational demand between network, while the accuracies are relatively close to each other. In fact, the ground truth in this application is rather noisy (smile is difficult to annotate as there are several levels between a full smile and no smile), and therefore the significance of the observed differences in accuracy are questionable. Nevertheless, one can claim that the mobilenets excel in accuracy/complexity tradeoff, and the early architectures (such as the VGG) should be abandoned in a real time implementation.

In addition, we proposed an asynchronous software architecture, which enables integration of other detection modules in a straightforward manner. Our experiments show that the proposed NVidia Jetson system level design enabled us to achieve the average processing time of 27.3 frames per second, reaching real time performance. The integration of other detection pipelines (age detection, gender detection) will be one of the future directions in this research.

Another direction of future work is to integrate the smile detector into a software framework, such as PRUNE \cite{prune}, that takes care of data transfers and synchronization between CPU cores and the GPU. Adopting such a framework would enhance the modularity of the software and on the other hand allow concentrating the development effort on the algorithms instead of multiprocessing, as well as enable even higher gains in processing time on a platform specifically designed for streaming data processing.

\begin{table}
 \centering
\captionof{table}{Accuracy comparison of proposed network architecture (Mobilenet ($\alpha=1, \rho=1$)) to the state-of-the-art method.}\label{tab:comparison}
\begin{minipage}{0.4\textwidth}
\resizebox{\columnwidth}{!}{
\begin{tabular}{l || c c} 
  & GENKI-4K & CelebA\\
  \hline
  \hline 
  SmileNet & 95.76\% & 92.81\% \\
  Proposed model &  93.6\% &  88.5\% \\
\end{tabular}
}
\end{minipage}
\end{table}

According to our smile detector use--case, the proposed approach can provide almost even three times faster average processing times on GPU than the reported state--off--the--art smile detectors. Besides, the smile detection accuracy of our method is close to that of the state--off--the--art methods.

\end{document}